\documentclass[%
 aapm,
 mph,%
 amsmath,amssymb,
 reprint,%
]{revtex4-2}

\usepackage{graphicx}
\usepackage{dcolumn}
\usepackage{bm}
\usepackage{makecell}
\usepackage[flushleft]{threeparttable}
\usepackage[mathlines]{lineno}
\usepackage{multirow}
\modulolinenumbers[5]

\begin{document}

\preprint{AAPM/123-QED}

\title[Ye et al.: Heterogeneous Decentralized Machine Unlearning with Seed Model Distillation]{Heterogeneous Decentralized Machine Unlearning with Seed Model Distillation}

\author{Guanhua Ye}
 \affiliation{The University of Queensland, Australia. E-mail: g.ye@uq.edu.au.}
 \author{Tong Chen}
 \affiliation{The University of Queensland, Australia. E-mail:  tong.chen@uq.edu.au.}
 \author{Quoc Viet Hung Nguyen}
 \affiliation{Griffith University, Australia. E-mail: henry.nguyen@griffith.edu.au.}
  \author{Hongzhi Yin (Corresponding Author)}
 \affiliation{The University of Queensland, Australia. E-mail: h.yin1@uq.edu.au.}
 


\begin{abstract}
As some recent information security legislation endowed users with unconditional rights to be forgotten by any trained machine learning model, personalized IoT service providers have to put unlearning functionality into their consideration. The most straightforward method to unlearn users' contribution is to retrain the model from the initial state, which is not realistic in high throughput applications with frequent unlearning requests. Though some machine unlearning frameworks have been proposed to speed up the retraining process, they fail to match decentralized learning scenarios. In this paper, we design a decentralized unlearning framework called HDUS, which uses distilled seed models to construct erasable ensembles for all clients. Moreover, the framework is compatible with heterogeneous on-device models, representing stronger scalability in real-world applications. Extensive experiments on three real-world datasets show that our HDUS achieves state-of-the-art performance.
\end{abstract}

\maketitle

\section{\label{sec:intro}Introduction}
The surge of edge computing and big data brings people personalized services in various domains like product recommendation \cite{imran2022refrs} and personalized healthcare analysis \cite{ye2022personalized}. In those services, users' edge devices (e.g., smartphones and smartwatches) play an important role in collecting user data and generating analytical feedback \cite{asare2021predicting, harirchian2022ml, chae2020development}, while providing a security and timeliness advantage compared with the outgoing centralized services. 
With an irreversible trend of decentralization where user data is treated as a non-shareable asset, enhancing collaborations between on-device machine learning models becomes the key to a high-performance, flexible, and privacy-preserving distributed learning framework.

Federated learning (FL) \cite{yang2019federated} and fully decentralized learning (FDL) \cite{1638541} are two representative distributed learning paradigms. A typical FL architecture is composed of multiple user clients and a central server, where the clients individually perform model updates based on their local data and the server gathers all local updates (e.g., by taking the average of submitted gradients) and then synchronizes all client models \cite{hsieh2020non}. 
Compared with depending on one centralized model, the use of local client models shortens the response time and lowers the risk of sensitive data leakage \cite{kasyap2021privacy}. However, FL still relies on a trusted central server to coordinate all clients throughout model training, which is not always guaranteed in practice. In contrast to FL, FDL frameworks train local models by allowing clients to directly exchange knowledge with neighbors in the communication network \cite{kempe2003gossip}, which bypasses the need for a central server. 


In both FL and FDL, device-wise collaborations are enabled by sharing local model parameters (e.g., weights or gradient) with the server or neighbors. It requires all clients to maintain a homogeneous model structure, such that knowledge can be shared across clients via equidimensional aggregation operations over different models' parameters \cite{bistritz2020distributed, long2023decentralized, qu2023semi, long2023model}. However, in real-world applications, users are more likely to possess devices with various hardware configurations, hence requiring different model structures for optimized performance \cite{bistritz2020distributed, chen2021learning, cai2020once}. For most of the existing distributed learning paradigms that only support collaborations among homogeneous client models, this is a fatal disadvantage in data-driven learning that heavily hinders flexibility and generalizability. This has been the key driver of enabling heterogeneous collaboration in distributed learning \cite{bistritz2020distributed, nguyen2017argument, hung2017computing}, where the key is to replace model parameter sharing with knowledge co-distillation between client models \cite{seo2020federated} via a publicly shared reference dataset. Specifically, each client's local model produces its own prediction on the reference dataset, commonly represented as logits in the prediction layer. As such, the implicit knowledge carried by the logits \cite{yuan2020revisiting} can be used to train a performant global model in FL \cite{zhang2021adversarial}, or to improve every local model by contrasting the logits between local and neighbor client models in FDL \cite{bistritz2020distributed}.


Meanwhile, as personalized algorithms are highly reliant on sensitive user data, a higher privacy standard is raised to protect user rights beyond decentralized learning paradigms. In recent information security legislations like the General Data Protection Regulation (GDPR) \cite{voigt2017eu} and the California Consumer Privacy Act (CCPA) \cite{pardau2018california}, a user's unconditional rights to be forgotten have been highlighted. Specifically, when a user quits any services, service providers should be able to not only delete the data collected from the user, but also fully remove her contribution to the learned machine learning models upon request \cite{shastri2019seven}.  Also, the rise of poisoning attacks to distributed learning frameworks \cite{bhagoji2019analyzing} further amplifies the need for unlearning clients with malicious or low-quality data to ensure robustness. Unfortunately, most existing distributed learning paradigms only allow users to deposit contributions to the global model without an option to withdraw. In this case, service providers will have to retrain a model from the ground up with the target user's data deleted, which incurs prohibitive time and resource consumption. 

\begin{figure*}[t]
\centering 
\includegraphics[scale=0.2]{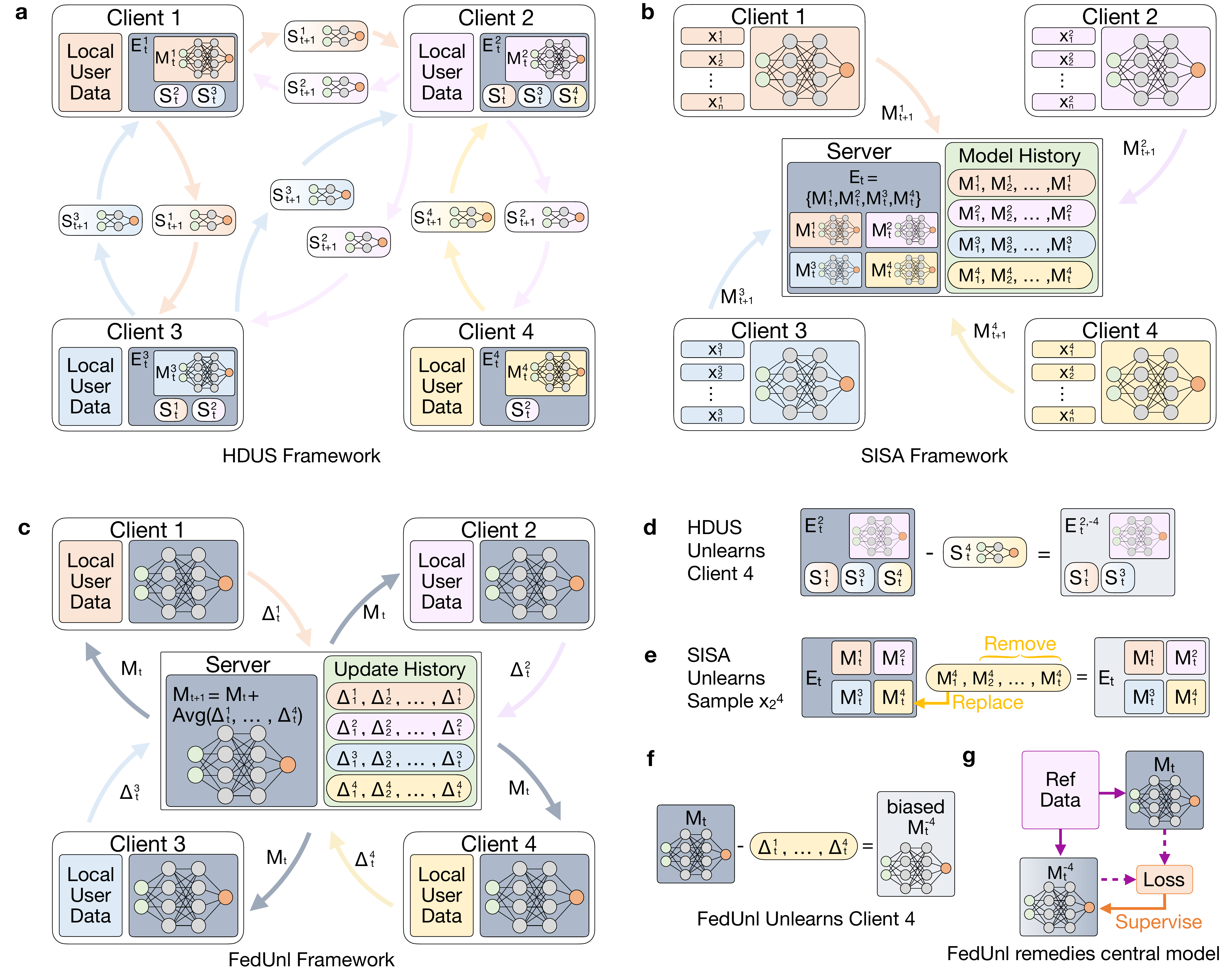}
\caption{Comparison of different unlearning frameworks. Note that four clients are used for demonstration purposes. \textbf{(a)} Our fully decentralized HDUS framework. Each local main model $M^i$ updates its parameters based on its local data and then supervises its seed model $s^i$ via a shared reference dataset. The received seed models collaboratively participate in the local final decision ($E^i$(x)) without introducing external information into the local main model ($M^i$). \textbf{(b)} Federated SISA framework. Clients train local models (with batch size = 1) based on the local data and then send each updated model to the server. In this way, each sample $x_t$ corresponds to one unique model $M_t$. The up-to-date models from all clients form an ensemble on the server for the final decision. \textbf{(c)} Federated FedUnl framework. Clients receive a central model from the server and train it based on local data. The server collects and stores all the updates to generate a new central model. \textbf{(d)} When client $4$ quits, HDUS removes its seed model from all recipients. \textbf{(e)} SISA does not support client-wise unlearning. Assuming the second sample from client $4$ is requested to be unlearned, the SISA server will replace client $4$'s latest model $M_t^4$ with the model before the second sample has appeared ($M_1^4$). \textbf{(f)} When client $4$ quits, FedUnl subtracts all its contributions from the central model. \textbf{(g)} After that, the unlearned central model distills knowledge from the previous one (before subtraction) to regain performance.} 
\label{Fig:Overview} 
\vspace{-0.5cm}
\end{figure*}

In this regard, some fast distributed unlearning techniques have been proposed to avoid a complete retraining cycle. SISA \cite{bourtoule2021machine} proposes to divide the data into multiple shards, and deploy a client model on each shard. Then on each shard, the model is trained in an incremental way (Fig~\ref{Fig:Overview}.b) with dynamically collected data samples, and a copy is saved for every checkpoint. The server will host and parallelly ensemble all learned client models. It has to be mentioned that SISA is a \textbf{sample-wise} unlearning framework, i.e., each unlearning request applies to only one data sample on the client. When the system receives an unlearn request, only the client hosting that data sample will restore its model to the checkpoint before this data sample is collected (Fig~\ref{Fig:Overview}.e), and then retrain the model with the updated data. Though SISA is an exact unlearning method that guarantees the elimination of all information about a data sample, it lacks the practicality for large-scale applications because of the need for keeping a copy of the model parameters for every checkpoint, which is inefficient for storage and infeasible in high throughput data streams. In contrast, \cite{wu2022federated} designs a federated unlearning (FedUnl) framework that stores the contributions of each client to the central model (Fig~\ref{Fig:Overview}.c). FedUnl is a \textbf{client-wise} unlearning approach, which can subtract an arbitrary client's contribution from the global model (Fig~\ref{Fig:Overview}.f), and remedy its performance by distilling knowledge from the full global model with a global reference dataset (Fig~\ref{Fig:Overview}.g). However, as the knowledge distillation inevitably introduces the target user's information back to the unlearned global model, it is categorized as an approximate unlearning method that provides a suboptimal privacy guarantee. Additionally, unlike SISA which takes a parallel ensemble of different models, FedUnl has to use a homogeneous model architecture across all clients, which contradicts the necessity of allowing heterogeneous model structures in distributed learning. It may seem that batch sample unlearning could serve as a potential solution to the problem of client-wise unlearning. However, in practice, sample-wise exact unlearning techniques necessitate a rollback of the model to a state where the target sample had not yet appeared. In the context of decentralized learning, this would mean rewinding the model to a state before the client joined the framework. As such, for any client that has joined the framework for a considerable amount of time, the framework needs to be retrained from a very early stage, which is impractical for real applications.

In light of this, we propose a  \textbf{H}eterogeneous \textbf{D}ecentralized  \textbf{U}nlearning framework with \textbf{S}eed model distillation (\textbf{HDUS}), which is designed for fast, memory-efficient, and exact unlearning that supports heterogeneous collaborative distributed learning. 
HDUS is an instance of the FDL paradigm, and an overview is presented in Fig.\ref{Fig:Overview}.a. In HDUS, each client owns: (1) its unique local dataset; (2) a reference dataset shared across all clients; (3) a main model trained with local data; and (4) a group of lightweight seed models shared by neighboring clients. 
Notably, each seed model in a client is trained on its corresponding neighbor, facilitated by distilling knowledge from the neighbor's main model with the shared reference dataset. The reference dataset only needs to follow the format of local client data and contains no client-specific data points (e.g., constructed with simulation/public data), thus ensuring that the distilled seed model does not reflect any personal and sensitive information.
The main model and seed models in the client constitute an ensemble model, so as to provide stronger performance and generalizability. 

Utilizing seed models, HDUS presents several advantages over existing distributed counterparts, as follows:

\begin{itemize}
\item HDUS enables direct information sharing between clients using seed models as secure intermediaries, significantly reducing dependency on central servers and achieving full decentralization. Rather than exchanging information from main models as in current heterogeneous distributed learning paradigms \cite{zhang2021adversarial,bistritz2020distributed}, knowledge transfer is facilitated by sharing distilled seed models across clients, which remain independent of their main models.
\item By circumventing traditional parameter sharing across client models, HDUS's design is compatible with heterogeneous model architectures, thereby maximizing usability within heterogeneous device networks.
\item The dissociative seed models serve as an add-on module comprising a neighbors' ensemble, allowing for straightforward removal of each neighbor's contribution for unlearning purposes, without necessitating retraining. Seed models are deliberately lightweight to ensure efficiency in memory usage and computation for clients.
\end{itemize}

To the best of our knowledge, this represents the first study aimed at devising an unlearning solution for decentralized collaborative learning with heterogeneous client models. As discussed in subsequent sections, our extensive experiments conducted on three real-world datasets demonstrate that HDUS is a highly versatile and effective framework in comparison with cutting-edge baselines.

\section{Related Work}

In this section, we analyze and summarize research backgrounds that are relevant to our work.

\subsection{Exact and Approximate Unlearning}
Machine learning is a field known for the development of algorithms capable of learning from data to make predictions or decisions \cite{goodfellow2016deep}. In contrast, unlearning is a more recent concept that has emerged as a response to the growing concerns about privacy and data security in machine learning applications \cite{voigt2017eu, pardau2018california}. Several approaches have been proposed to tackle the unlearning problem. One approach involves the use of deletion tokens, where each data point is associated with a unique token, allowing for efficient removal of the corresponding data \cite{foerster2017input}. Another approach is the use of selective amnesia, which is a technique to forget specific subsets of data without affecting the overall model \cite{aljundi2018memory}. An alternative approach to unlearning is to leverage the inherent properties of differentially private machine learning techniques, which provide a formal privacy guarantee by adding noise to the model during training \cite{abadi2016deep, dwork2014algorithmic}. 

Exact unlearning methods aim to completely remove a user's contribution from a learned model, providing a strong privacy guarantee. However, such methods often require the model to be retrained from scratch, which can be computationally expensive and time-consuming. One example of an exact unlearning approach is the leave-one-out retraining method, where the model is retrained using the entire dataset except for the user's data to be unlearned \cite{lee2018concentrated}. Another approach involves the use of selective influence estimators \cite{ghorbani2019data}, which can identify the influence of individual data points on the model parameters. Although effective in terms of privacy preservation, the high computational cost of these methods limits their practicality in real-world applications \cite{ene2017decomposable}.

In contrast, approximate unlearning methods attempt to remove a user's contribution from a learned model without requiring complete retraining. This reduces the time and computational resources needed for unlearning, but at the cost of potentially weaker privacy guarantees. One such approach is the gradient surgery method proposed by Cao and Yang \cite{cao2015towards}, which involves updating the model parameters using the negative gradient of the target user's data. Another example is \cite{nguyen2020variational}, which employs a probabilistic model to approximate the unlearning process. The use of elastic weight consolidation \cite{kirkpatrick2017overcoming} and model compression techniques \cite{hinton2015distilling} have also been explored to facilitate approximate unlearning. Although approximate unlearning is generally more efficient than exact unlearning, its compliance with strict privacy regulations remains to be an open question \cite{thudi2022unrolling}.

\subsection{Fully Decentralized Learning}

Fully decentralized learning (FDL) is a subfield of distributed learning that allows multiple clients to collaboratively learn models without relying on a central server. Consensus optimization has been widely used in FDL to facilitate coordination among clients, which aims to minimize the global objective function by reaching consensus on the model parameters among all clients \cite{lai2018decentralized}. A straightforward algorithm in this context is the decentralized gradient descent (DGD) method \cite{nedic2010convergence}, where local gradients are employed to update model parameters in isolation.

Another important aspect of FDL is its applicability to various machine learning tasks, such as classification, regression, and reinforcement learning. For example, decentralized consensus ADMM (Alternating Direction Method of Multipliers) has been applied to solve decentralized support vector machine (SVM) problems \cite{forero2010consensus}. In reinforcement learning, distributed Q-learning and actor-critic algorithms have also been proposed for FDL settings \cite{iftikhar2022ai}. To ensure privacy in FDL, various techniques have been proposed, such as secure multi-party computation (SMPC) \cite{goldreich1998secure}, which allows clients to perform computations on encrypted data without revealing the original data. Homomorphic encryption is another approach that enables clients to perform computations on encrypted data without the need for decryption \cite{gentry2009fully}. However, those methods come at a cost of excessive computational overheads, which are less favored by large-scale applications. 

\subsection{Ensemble Learning}
Ensemble learning is a technique that combines multiple learning models to improve the overall performance and generalization of the final model. This approach has been widely used in various machine learning applications, including classification, regression, and reinforcement learning \cite{zhou2012ensemble}. Some popular ensemble learning methods include bagging \cite{breiman1996bagging}, boosting \cite{schapire1990strength}, and stacking \cite{wolpert1992stacked}.

One of the key advantages of ensemble learning is its ability to reduce overfitting and improve the robustness of the final model \cite{ye2023heterogeneous}. This is achieved by aggregating the predictions of multiple base models, which are typically trained on different subsets of the data or using different algorithms \cite{dietterich2000ensemble}. Research has shown that ensembles of diverse models can often achieve better performance than any individual model, as they can effectively capture the strengths of each base model while mitigating their weaknesses \cite{rokach2010ensemble}.

In the context of the proposed HDUS framework, ensemble learning is employed to enable clients to effectively leverage the knowledge distilled from neighboring clients' seed models. This not only enhances the performance and generalizability of the ensemble model, but also allows for easy removal of a neighbor's contribution for unlearning purposes without the need for retraining. The incorporation of ensemble learning in decentralized and distributed settings has been explored in previous work, such as federated ensemble learning \cite{demertzis2023federated} and decentralized ensemble learning \cite{yu2019decentralized}, which further facilitates ensemble learning in the HDUS framework.

\section{Heterogeneous Decentralized Machine Unlearning}
We unfold the design of the HDUS framework in this section.

\begin{figure*}[t!] 
\centering 
\includegraphics[scale=0.25]{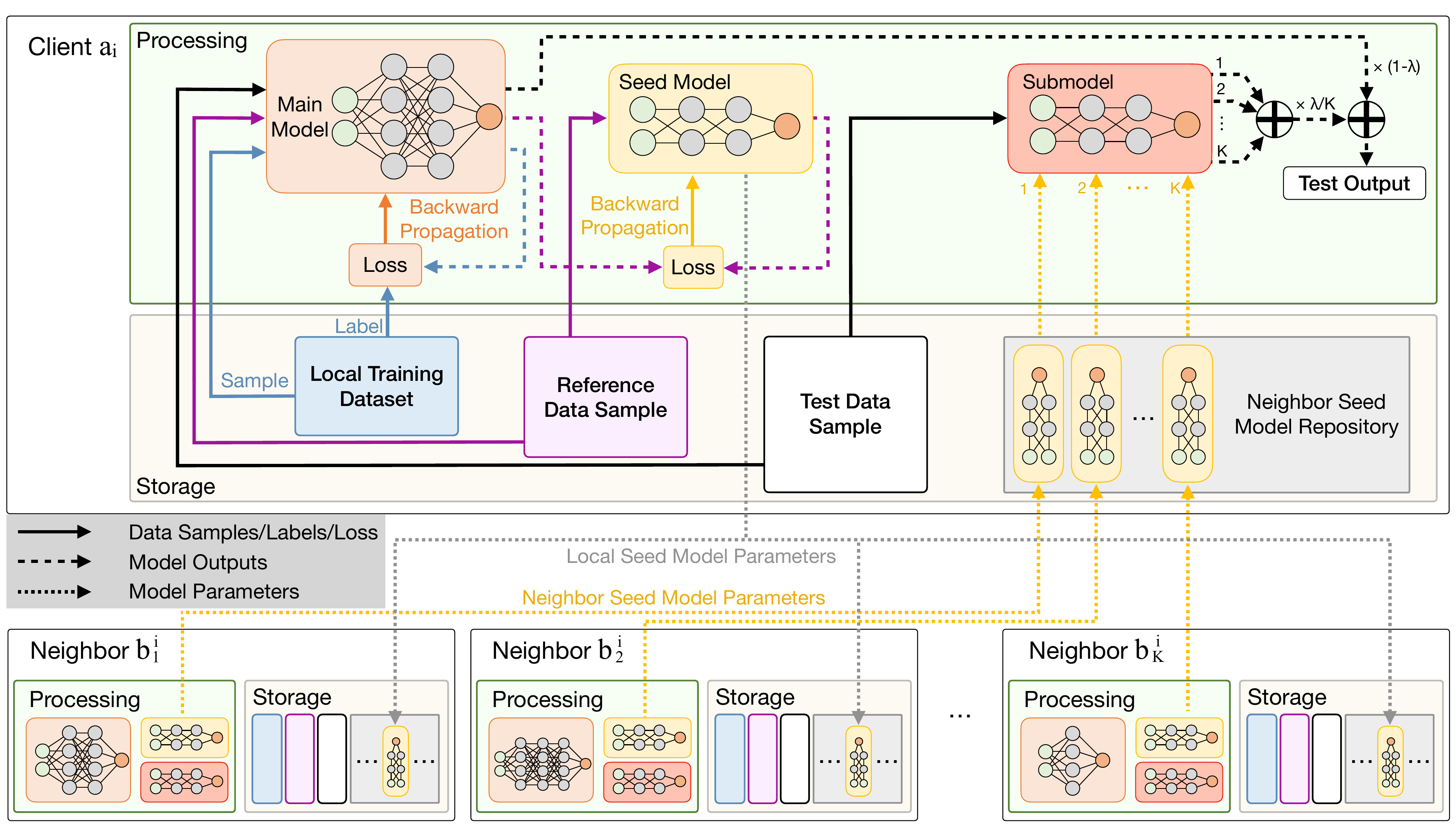}
\caption{The detailed framework of HDUS. For a user client $a_i$ in the network, it trains its main model (orange) with its local training dataset (blue) and a seed model (yellow) that distills knowledge from the main model. An unlabeled reference dataset (purple) is introduced to support the knowledge distillation. The collaboration between $a_i$ and its neighbor $b_1^i$ to $b_K^i$ is actualized by exchanging seed model parameters. The receiving parameters are stored in the seed model repository (grey). In the testing stage (black arrow), the seed models in the repository will form a submodel ensemble (red) to generate $K$ soft labels. The final decision is composed of the output from the main model and these soft labels. Note that the seed model uses a model structure that is different from the main model and it is isolated from the sensitive local training data. Meanwhile, the main model structure is heterogeneous across clients. } 
\label{fig:Model} 
\end{figure*}

\subsection{Preliminaries}
In a decentralized collaborative system with $N$ clients/users $\mathcal{A} = \{a_1, a_2, ..., a_N\}$, client $a_i$ possesses $M_i$ local training samples with $F$-dimensional features $X_i \in \mathbb{R}^{ M_i \times F}$ and their one-hot labels $Y_i \in \mathbb{R}^{M_i \times C}$ over $C$ classes. We name $\mathcal{D}_i^{loc} = \{X_i, Y_i\}$ as the local dataset of client $a_i$. For each $a_i \in \mathcal{A}$, it can train a local deep neural network (DNN) $f(\theta_i,\cdot)$ with parameterization $\theta_i$ by minimizing the loss $\ell(f(\theta_i, X_i), Y_i)$. Assuming $a_i$ has identified $K_i$ neighboring clients $\mathcal{B}_i = \{b_1^i, b_2^i, ..., b_K^i\}$ ($\mathcal{B} \in \mathcal{A}, a_i \notin \mathcal{B}$) and $\mathcal{S}_i = \{s_1^i, s_2^i, ..., s_K^i\}$ represents the knowledge/information (e.g., model parameters in FL) from those neighbors, then the collaboratively learned model for $a_i$ is denoted by $f(\theta_i',\cdot) = F(f(\theta_i,\cdot), \mathcal{S}_i)$, where $\theta_i'$ is updated model parameters that integrates knowledge from $\mathcal{S}_i$. Intuitively, $f(\theta_i',\cdot)$ performs better than $f(\theta_i,\cdot)$. When a client, say $b_j^i \in \mathcal{B}_i$ quits the system, in addition to deleting its own model $f(\theta_j,\cdot)$ and data $X_j$, its information also needs to be unlearned by $f(\theta_i',\cdot)$ for all $a_i$ having $b_j^i \in \mathcal{B}_i$. Specifically, client $a_i$ needs to adjust its model $R(f(\theta_i',\cdot), s_j^i) = F(f(\theta_i,\cdot), \mathcal{S}_i - s_j^i)$ by removing $b_j^i$'s contribution as if $b_j^i$ never participated in the learning process. For this purpose, we aim to design an exact unlearning framework that can fully eliminate the influence of $b_j^i$ while keeping the services in all other clients uninterrupted.

\subsection{HUDS framework}
Fig.\ref{fig:Model} provides a complementary graphical view of HDUS from a single client's perspective. 
In the training phase (blue arrow in Fig.\ref{fig:Model}), the local main model $f(\theta_i,\cdot)$ updates its parameters based on the local training data $X_i$. 
To avoid privacy breaches, user data $X_i$ and local model parameter $\theta_i$ cannot be shared during collaborative learning. 
Therefore, we build a new communication pathway by introducing a shared reference dataset $X_i^{ref}$ in every client $a_i$ to train a seed model $\theta_i^{seed}$. Instead of designing supervised tasks to learn $\theta_i^{seed}$, the seed model $f(\theta_i^{seed},\cdot)$ distills knowledge from the main model $f(\theta_i,\cdot)$ by minimizing:
\begin{align}\label{eq:distill}
\mathcal{L}^*=\min_{\theta_i^{seed}} \mathcal{L}\bigg(f(\theta_i^{seed}, X_i^{ref}),  \sigma_{T} ( f(\theta_i, X_i^{ref}))\bigg),
\end{align}
where the $\mathcal{L}$ is some loss function like Kullback–Leibler (KL) divergence. We term this process the incubating phase (purple arrow in Fig.\ref{fig:Model}). $\sigma_T(\cdot)$ is a modified softmax function with a temperature parameter $T$ to control the strength of distillation \cite{papernot2016distillation}. Specifically, for a model output vector $y = \{d_1, d_2,\cdots d_C\}$, we define $\sigma_T(y) = \{s(d_1,T),s(d_2,T),\cdots,s(d_C,T)\}$ where $s(d_i,T) = \frac{\exp(\frac{d_i}{T})}{\sum_{i=1}^C\exp(\frac{d_i}{T})}$. 
Notably, each client aims to align its seed model as closely as possible with its main model, and therefore,  $X_i^{ref}$ is neither required to be labelled nor identical across all clients. This makes the construction of reference datasets more convenient and feasible for various decentralized applications, e.g., each client can use a simulated dataset in healthcare.
Furthermore, since $f(\theta_i^{seed},\cdot)$ is not associated with $\mathcal{D}_i^{loc}$ during training, a user's personal data is highly unlikely to be restored from $\theta_i^{seed}$.

The seed model is smaller than the main model while being able to mimic its decision-making behaviors, which provides a lightweight yet more secure communication protocol. During collaborative learning, client $a_i$ will send its seed model parameters $\theta_i^{seed}$ to all neighbors $\mathcal{B}_i$ and receive $K$ seed models $\mathcal{S}_i$ where $s_j^i = \theta_i^{seed}$ for $b_j^i \in \mathcal{B}_i$. These model parameters are stored in a dedicated repository of $a_i$. Then in the test/inference phase (black arrow), each client generates an ensemble model output for an unseen data point $x$ via:
\begin{eqnarray}\label{eq:learning_obj}
F(f(\theta_i, x), \mathcal{S}_i) = (1-\lambda) f(\theta_i, x)+ \frac{\lambda}{K}\sum_{k=1}^{K} f(s_k^i,x),
\end{eqnarray}
where $\lambda$ is a trade-off hyperparameter adjusting the contribution from all $K$ neighbors' seed models. Note that the local seed model $\theta_i^{seed}$ is not involved in Eq.(\ref{eq:learning_obj}) as the main model is essentially its stronger variant.

\subsection{Handling Unlearning Requests} 
In conventional weight/gradient-based decentralized learning frameworks, the knowledge from neighbors influences the main model $f(\theta_i',\cdot)$ throughout the whole training process. Therefore, when $b_j^i$ quits $\mathcal{B}_i$ (and $\mathcal{A}$) and requests its information $s_i^j$ (i.e., model parameters in conventional frameworks) to be unlearned by $f(\theta_i',\cdot)$, client $a_i$ needs to restore its model to the initial state before $s_i^j$ was involved. 
Moreover, client $a_i$ will spread its model parameters (e.g., $\theta_i'$ and $\Delta\theta_i'$) to other clients. In the worst case, all clients in the network have to restore their main models, which introduces a prohibitive cost in model retraining. 
In contrast, in our proposed framework HDUS, the knowledge from neighbors is never fused with $a_i$'s main model. Specifically, each received seed model is treated as a sub-model in $a_i$'s ensemble, which allows for convenient machine unlearning while giving $a_i$'s main model a performance boost. In this way, upon receiving an unlearning request from neighbor $b_j^i$ with seed model $s_j^i$, the updated model ensemble for $a_i$ is:
\begin{align}\label{eq:unlearning_obj}
&F(f(\theta_i,x), \mathcal{S}_i-s_j^i)\nonumber \\
= &(1-\lambda) f(\theta_i, x) + \frac{\lambda}{K-1}\sum_{k=1, k\neq j}^{K} f(s_k^i,x),
\end{align}
which essentially reflects the process of removing $b_j^i$ from $\mathcal{B}_i$ and thus removing $s_j^i$ from $\mathcal{S}_i$. Since all the knowledge from neighbor $b_j^i$ is contained and only contained by $s_j^i$, Eq.(\ref{eq:unlearning_obj}) can be regarded as an exact unlearning process. In other words, the remaining elements in the ensemble can be put into use without any modification right after the unlearning operation.

\section{Experiments} \label{sec:result}

\begin{table*}[t]
\normalsize
\centering
\caption{Comparison of different methods in decentralized on-device unlearning.}
\vspace{-0.4cm}
\begin{center}
\begin{threeparttable}
\begin{tabular}{cccccccc}
\hline
\multirow{3}{*}{}
& \makecell{Exact\\ Unlearning}
& \makecell{Performance \\ Recovery\\  Acceleration\tnote{1}}
& \makecell{Uninterrupted \\Service After\\Unlearning\tnote{2}}
& \makecell{Client-wise \\ Unlearning}
& \makecell{Without \\ Storing \\ Model History}
& \makecell{Without \\ Central \\ Server}  
& \makecell{Heterogeneous \\ Model \\ Structure}
  \\  
\hline
ISGD & × & × & × & × & × & \checkmark & \checkmark \\
SISA  & \checkmark & \checkmark & \checkmark & × & × & × & ×  \\   
FedUnl  & × & \checkmark & × & \checkmark & × & × & ×   \\
DSGD & \checkmark & × & × & \checkmark & \checkmark & \checkmark & ×   \\ 
\textbf{HDUS} & \checkmark & \checkmark & \checkmark & \checkmark& \checkmark & \checkmark & \checkmark   \\  
\hline 
\end{tabular}
\begin{tablenotes}
\footnotesize
\item \checkmark\  Supported  \ \ \ \ \ \ \ \ \ \ \ \  ×\ \  Not supported or unavailable
\item[1] In comparison with retraining the model from its initial state. 
\item[2] All remaining clients possess a usable model (ensemble) before unlearning finishes.
\end{tablenotes}
\end{threeparttable}
\label{tab:Method_Comparision}
\end{center}
\vspace{-1.0cm}
\end{table*}

\begin{table}[t]
\caption{Characteristics of datasets.}
\vspace{-0.3cm}
\setlength\tabcolsep{2.2pt}
\begin{center}
\begin{tabular}{cccccc}
\hline
Dataset &  \makecell[c]{Input Size} & \makecell[c]{\#Training} & \makecell[c]{\#Test } & \makecell[c]{\#Reference} & \makecell[c]{\#Class}  \\
\hline
MNIST & $28\times28$ & $48,000$ & $12,000$ & $10,000$ & $10$  \\  
FMNIST & $28\times28$ & $48,000$ & $12,000$ & $10,000$& $10$  \\  
Cifar10 & $30\times30\times3$ &  $40,000$  & $10,000$ & $10,000$ & $10$ \\
\hline
\end{tabular}
\label{tab:Dataset}
\end{center}
\vspace{-1.0cm}
\end{table}

This section presents our findings about the proposed HDUS framework via comparative analyses and experiments.

\subsection{Comparative Analysis on Functionality}
We discuss and compare the functionality of our approach with four representative baselines that support both learning and unlearning in a distributed environment:
\begin{itemize}
\item \textbf{ISGD}~\cite{lian2017can}: A naive framework where isolated stochastic gradient descent (ISGD) assumes each client trains its own model independently, which puts all local models at the risk of overfitting and underperforming when local training data is insufficient. Unlearning a client's information simply corresponds to deleting this client's model and data.
\item \textbf{SISA}~\cite{bourtoule2021machine}: A sample-wise unlearning approach designed for FL. Each client holds a unique part of the full dataset and trains a local model in an incremental, instance-by-instance way. The central server then collects all client models and aggregates their outputs during inference. For each local model, SISA stores $T$ historical states (checkpoints) for $T$ data slices. In sample-wise unlearning, suppose the target data sample is in the $t$-th ($t\leq T$) slice, the client only needs to retrain the local model from the checkpoint $t-1$ rather than the initial state.
\item \textbf{FedUnl}~\cite{wu2022federated}: A variant of FedAvg \cite{mcmahan2017communication} that enables client-wise unlearning. In FedUnl, each client receives a copy of the global model from the central server, and updates the model based on its local dataset. The central server then collects all the updated model weights from the clients, calculates an average model, and distributes the new global model back to all clients. 
For unlearning an entire client's information, FedUnl further requires the central server to store historical updates from all clients. Whenever one client quits, the central server will subtract all its updates from the global model. As a remedy for the potentially biased global after unlearning, additional knowledge distillation from the original model is introduced, which consequently brings back implicit knowledge about the quitting client and makes FedUnl an inexact/approximate unlearning method.
\item \textbf{DSGD}~\cite{koloskova2020unified}: A homogeneous FDL framework based on decentralized SGD, where each client has a unique dataset and a local model with the same architecture. Without any central servers, the clients directly communicate with their neighbors via model weight sharing. When a client quits, all remaining clients need to retrain their models from the initial state. 
\end{itemize}

Table~\ref{tab:Method_Comparision} gives a functional comparison between our method and the four baselines. While HDUS ticks all the boxes, we briefly analyze the functional differences of all baselines. ISGD is the simplest framework where all clients are trained in silos without any connection. Since there is no information dissemination, other clients do not need to respond to any unlearning request. SISA and FedUnl are two federated unlearning methods. Both of them need to store the model's historical updates for unlearning, bringing immeasurable storage demand for large-scale applications. Besides, the sample-wise unlearning design in SISA makes it struggle to scale to client-wise unlearning. FedUnl only supports inexact client-wise unlearning, which offers a suboptimal privacy guarantee and dissatisfies some latest legislations. DSGD is an FDL framework that unlearns a client by retraining its primordial model. It is worth noting that, among all baseline frameworks, only ISGD supports learning heterogeneous models across clients, and only SISA can keep its service uninterrupted during the unlearning process (i.e., no model retraining needed for any involving clients), which are two critical factors associated with the real-life applicability of distributed learning systems.  

\begin{table}[b]
\normalsize
\caption{Client model allocations in both settings.}
\vspace{-0.4cm}
\begin{center}
\begin{tabular}{ccccc}
\hline
 Setting & Dataset & \#Small & \#Medium & \#Large  \\
\hline
 & MNIST & $0$ & $0$ & $6$  \\  
Homogeneous & FMNIST & $0$ & $0$ & $6$  \\  
& Cifar10 & $0$ & $0$ & $5$ \\
\hline
& MNIST & $2$ & $2$ & $2$  \\  
Heterogeneous & FMNIST & $2$ & $2$ & $2$  \\  
& Cifar10 & $1$ & $2$ & $2$ \\
\hline
\end{tabular}
\label{tab:Configuration}
\end{center}
\vspace{-1.0cm}
\end{table}

\begin{table}[b!]
\caption{Sizes of different models in MB.}
\vspace{-0.4cm}
\begin{center}
\begin{tabular}{c|cc|cc}
\hline
Type & Model & Size (MB) & Model & Size (MB)  \\
\hline
Small & ResNet8 & $0.015$ & MobileNet-S & $0.022$ \\  
Medium & ResNet18 & $1.044$ & MobileNet-M & $1.362$\\  
Large & ResNet50 & $2.907$ & MobileNet-L & $8.665$ \\
\hline
\end{tabular}
\label{tab:Model_Size}
\end{center}
\vspace{-0cm}
\end{table}

\subsection{Learning Effectiveness}

\begin{table*}[t!]
\small
\caption{Classification accuracy of all frameworks in both homogeneous and heterogeneous settings.}
\vspace{-0.6cm}
\begin{center}
\begin{tabular}{c|c|c|c|c|c|c|c}
\hline
\multirow{2}{*}{Base Model} & \multirow{2}{*}{Framework} & \multicolumn{3}{c|}{Homogeneous Setting} & \multicolumn{3}{c}{Heterogeneous Setting} \\ 
\cline{3-8}
 & & MNIST & FMNIST &  Cifar10  & MNIST & FMNIST & Cifar10 \\
\hline
& ISGD 
& $0.9872\pm 0.0018$ 
& $0.9034\pm 0.0076$
& $0.7026\pm 0.0092$
& $0.9828\pm 0.0022$
& $0.8796\pm 0.0054$
& $0.6563\pm 0.0083$\\
& SISA-A
& $0.9913\pm 0.0022$ 
& $0.9112\pm 0.0054$ 
& $0.7408\pm 0.0167$
& $0.9822\pm 0.0031$ 
& $0.8660\pm 0.0068$ 
& $0.5190\pm 0.0351$\\  
ResNet & FedUnl 
& $0.9929\pm0.0011$ 
& $\textbf{0.9254}\pm0.0091$ 
& $\textbf{0.7834}\pm0.0075$ 
& $0.9825\pm0.0026$ 
& $0.8871\pm0.0091$ 
& $0.6204\pm0.0180$\\  
& DSGD 
& $\textbf{0.9932}\pm0.0017$ 
& $0.9185\pm0.0050$ 
& $0.7236\pm0.0083$ 
& $0.9816\pm0.0012$ 
& $0.8637\pm0.0069$ 
& $0.5039\pm0.0229$\\
& HDUS 
& $0.9910\pm0.0006$ 
& $0.9118\pm0.0032$ 
& $0.7482\pm0.0094$ 
& $\textbf{0.9856}\pm0.0026$ 
& $\textbf{0.8884}\pm0.0050$ 
& $\textbf{0.6808}\pm0.0097$\\
\hline
\hline
& ISGD 
& $0.9851\pm 0.0016$
& $0.9012\pm 0.0027$
& $0.7027\pm 0.0083$
& $0.9829\pm 0.0042$
& $0.8825\pm0.0021$
& $0.6637\pm 0.0101$\\
& SISA-A
& $0.9914\pm 0.0021$ 
& $0.9140\pm 0.0039$ 
& $0.7492\pm 0.0081$
& $0.9813\pm 0.0021$ 
& $0.8673\pm 0.0055$ 
& $0.5217\pm 0.0092$
\\
MobileNet & FedUnl 
& $\textbf{0.9945}\pm 0.0012$ 
& $\textbf{0.9268}\pm 0.0021$ 
& $\textbf{0.7915}\pm 0.0032$ 
& $0.9825 \pm 0.0017$ 
& $0.8878\pm 0.0057$ 
& $0.6283\pm 0.0129$
\\
& DSGD 
& $0.9940\pm 0.0015 $ 
& $0.9162\pm 0.0027$ 
& $0.7330\pm 0.0085$ 
& $0.9817\pm 0.0036$ 
& $0.8890\pm 0.0078$ 
& $0.5136\pm 0.0178$
\\
& HDUS 
& $0.9900\pm 0.0014$ 
& $0.9100\pm 0.0029$ 
& $0.7528\pm 0.0060$ 
& $\textbf{0.9875}\pm 0.0044$ 
& $\textbf{0.9023}\pm 0.0049$ 
& $\textbf{0.7152}\pm 0.0095$\\
\hline
\end{tabular}
\label{tab:Classification}
\end{center}
\vspace{-0.3cm}
\end{table*}

\begin{figure*}[t!] 
\centering 
\includegraphics[scale=0.375]{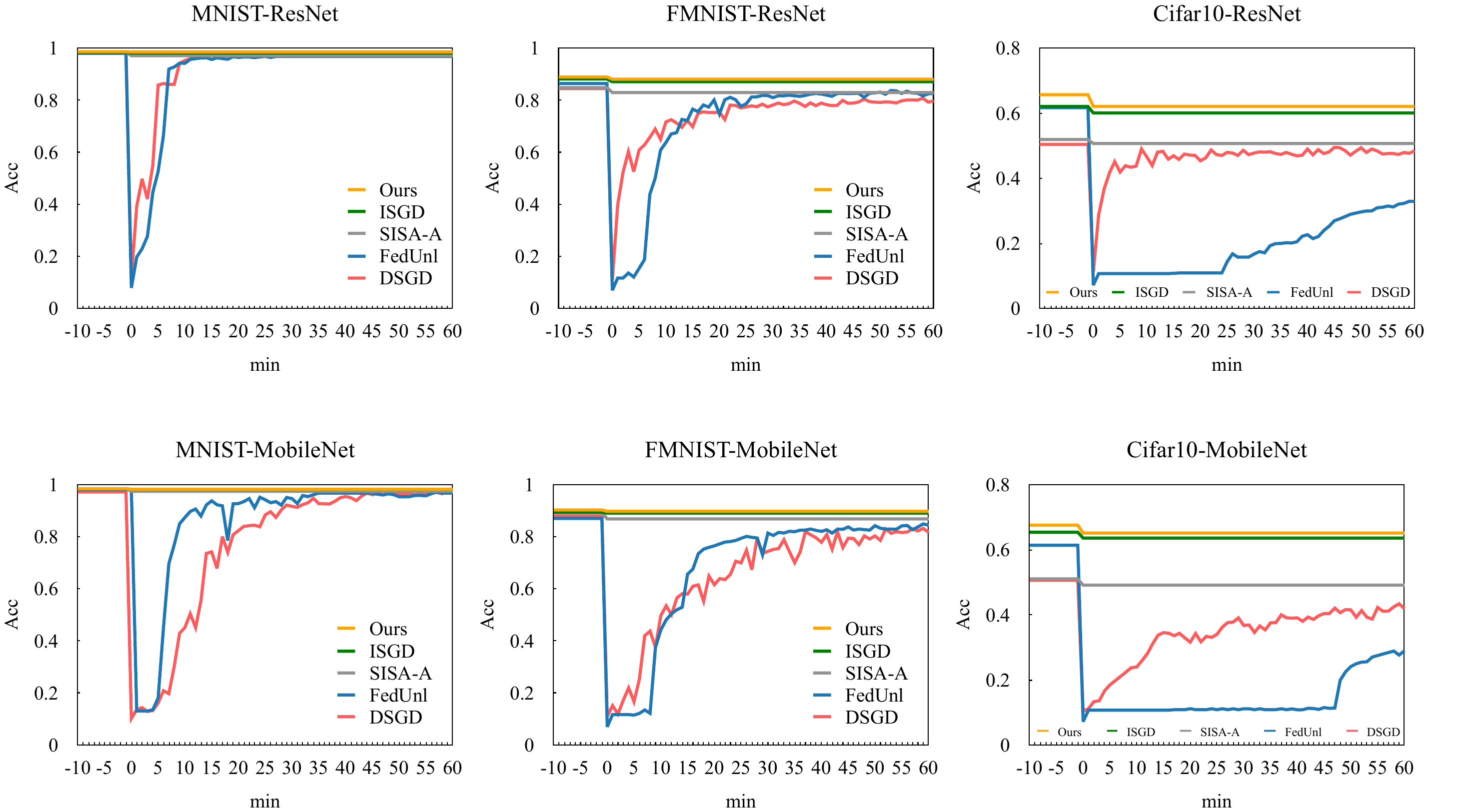}
\vspace{-1cm}
\caption{The unlearning process of different frameworks in the heterogeneous setting. All frameworks unlearn one client from $t=0$. The time consumption of deleting the corresponding (seed) model in HDUS, SISA-A, and ISGD is lower than $1$ minute. Runtime is measured on a single RTX 3090 Ti GPU.} 
\label{fig:Unlearning} 
\vspace{-0.5cm}
\end{figure*}

Next, we compare the performance of HDUS and four baselines. The experiments are conducted on three benchmark datasets commonly used in decentralized classification tasks, i.e., MNIST\cite{lecun1998mnist} (a greyscale image dataset for handwritten digits from $0$ to $9$), FMNIST\cite{xiao2017fashion} (a greyscale image dataset for 10 types of fashion items), and Cifar10\cite{krizhevsky2009learning} (a 10-class color image dataset). 
Detailed statistics of these three datasets are listed in Table~\ref{tab:Dataset}. The numbers of clients on all three datasets are 6, 6, and 5, respectively. All datasets have 10 distinct class labels. To assign clients with local data, we randomly draw an equal amount of non-overlapping instances from the training set, while each client only samples from a unique set of 9 classes to further simulate the non-I.I.D. nature of distributed datasets. For the determination of the optimal value for $|D^{ref}|$, an grid search was conducted, ranging from $4,000$ to $10,000$. The investigation unveiled that as $|D^{ref}|$ escalated, each seed model progressively aligned with the main model, thereby enhancing the overall framework performance. Consequently, for each data source, we reserve $10,000$ samples to use as a reference dataset. It is of note that these sample labels were exclusively provided to FedUnl upon request.

To make SISA compatible with client-wise unlearning, we modified SISA by treating each non-overlapping data shard as a user's personal dataset, and the ensemble of all local models as a central server model. We mark this amended version of SISA as SISA-A. 
For FDL frameworks (DSGD and HDUS), we let every client in FDL frameworks communicate with all other clients during learning/unlearning, so as to maintain a fair comparison with FL frameworks (SISA-A and FedUnl) that coordinate all clients via the global model. The homogeneous scenarios and heterogeneous scenarios are simulated by deploying different ResNet \cite{he2016deep} and MobileNetV2 \cite{howard2018inverted} versions on clients, as shown in Table~\ref{tab:Configuration}. In heterogeneous scenarios, MobileNet-L is the default model size, while MobileNet-M and MobileNet-S are manually pruned versions to approximate the size of ResNet8 and ResNet18, respectively. The detailed model sizes are listed in Table~\ref{tab:Model_Size}. Note that for baselines that do not support heterogeneous model communication (i.e., SISA-A, FedUnl, and DSGD), all clients are assigned the smallest model in heterogeneous scenarios to accommodate the lowest budget. 

We validate the classification effectiveness of HDUS and four baselines, where the average test accuracy and standard deviations of five independent runs are reported in Table~\ref{tab:Classification}. In the homogeneous setting, HDUS's performance is on par with different distributed learning frameworks. We also observe that HDUS is superior to SISA-A on FMNIST and Cifar10 under the homogeneous setting, which suggests that sharing a distilled small model is more cost-effective than sharing the entire local main model.  
When transferred to the heterogeneous setting, all methods are subject to a noticeable performance drop. Moreover, baselines with inter-client communication (i.e., SISA-A, FedUnl, and DSGD) perform even worse than naive ISGD in Cifar10. This is because the constraint on the total model parameter submerges the improvement from communication, especially in complex tasks. On the contrary, HDUS merely loses $1.25\%$ accuracy on average in heterogeneous scenarios, outperforming all baselines in the heterogeneous setting. The results also showcase the advantage of the collaborative learning protocol for heterogeneous models proposed in HDUS. While the HDUS framework is primarily evaluated on classification tasks in this work, the ensemble of knowledge from decentralized models can provide generalizability across a broad spectrum of tasks, such as ranking and regression, where non-I.I.D. data is also present.

\begin{figure*}[t!] 
\centering 
\includegraphics[scale=0.365]{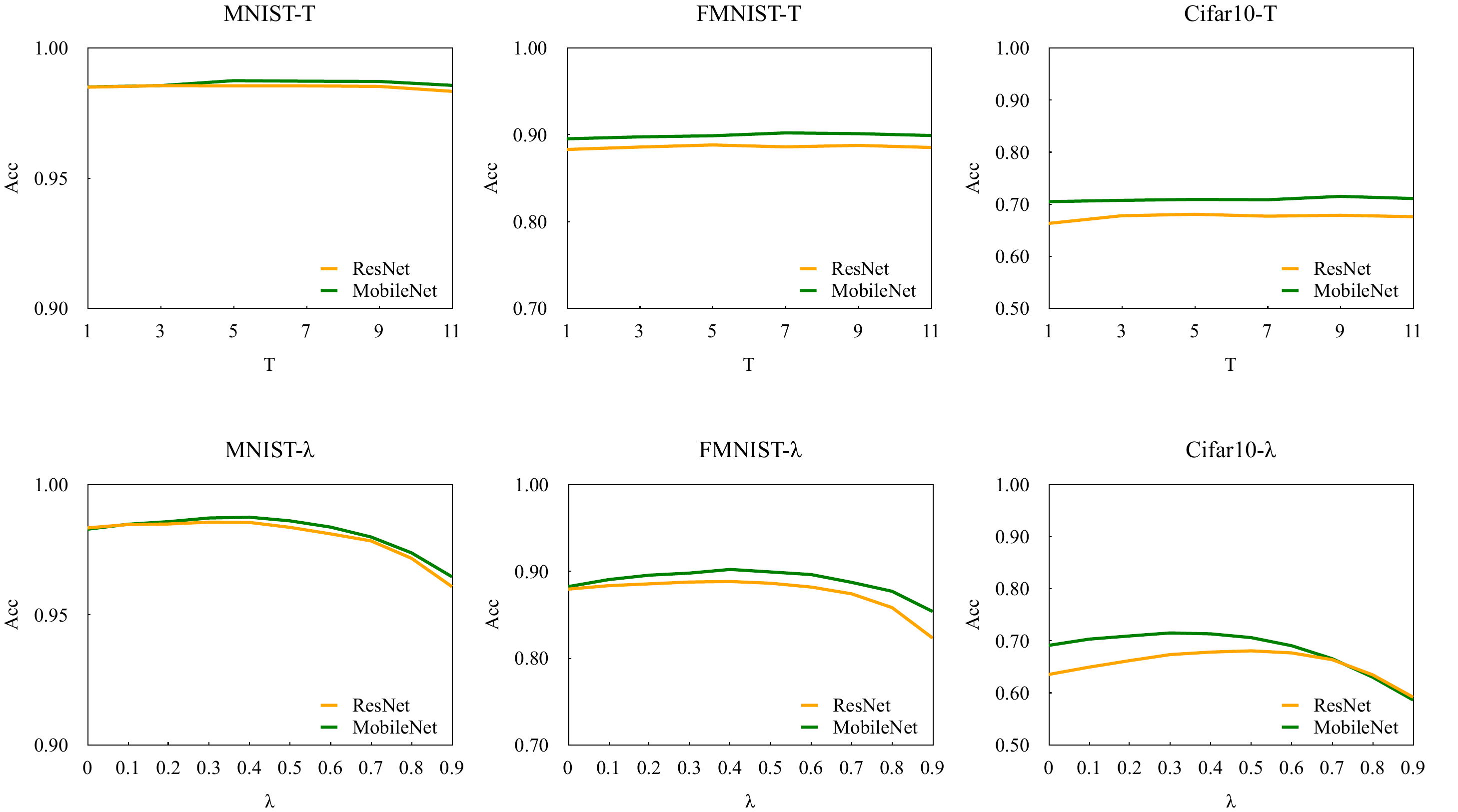}
\caption{The impact of different hyperparameter values in the heterogeneous setting, where the results are recorded by fixing all other hyperparameters and varying the value of $T$ or $\lambda$.} 
\label{fig:Hyper} 
\vspace{-0.4cm}
\end{figure*}

\subsection{Unlearning Effectiveness}
 We simulate the scenario that after all models are well-trained, one client in the framework sends an unlearning request to its neighbors (or the server) at time $t=0$. This means all remaining clients need to erase the quitting client's historical contributions from their local models while maintaining performance. The unlearning procedure is relatively straightforward for SISA-A and HDUS: since the knowledge from the quitting client is not fused into its neighbors' local models, they can simply adjust the ensemble (e.g., the global ensemble model in SISA-A and the seed model ensemble in HDUS) by removing the quitting client to achieve exact unlearning. 
 However, for FedUnl and DSGD, it will take much more effort to remove the footprint of the quitting client. As the knowledge of the quitting client has blended into all client models in the learning process, these frameworks have to execute a complicated inexact unlearning procedure (FedUnl) or retrain all models from the ground up (DSGD). 
 
 The test accuracy curves of all methods in the heterogeneous setting are presented in Fig.\ref{fig:Unlearning}. All unlearning processes are performed on the same GeForce RTX 3090 Ti GPU to allow for fair runtime measurement. When $t<0$, the test accuracy is evaluated on all clients, and when $t\geq0$, it is evaluated on all but the quitting clients. As the results demonstrate, HDUS outperforms all baselines consistently across datasets and model selections. 
Principally, ensemble frameworks (HDUS and SISA-A) can function seamlessly during unlearning, showcasing their potential in real-life scenarios with highly frequent unlearning requests. On the contrary, FedUnl and DSGD need much more time to return to their original performance level. The retraining/distillation process will be shortened if stronger computing power is available, but a service interruption is still unavoidable. If the unlearning requests come one after another (and even collide with the unfinished unlearning process), the usability of FedUnl and DSGD will degrade further. 
Besides, we notice that DSGD recovers faster than FedUnl, especially on the most complicated Cifar10 dataset. The reason is that models in FedUnl have two optimization objectives: (1) the classification error on each local dataset and (2) the discrepancy from the global model before the unlearning operation. Consequently, FedUnl's convergence rate is slower than DSGD's. However, given enough time, FedUnl can theoretically surpass DSGD in performance.
Another observation is that both DSGD and FedUnl recover slower when deploying MobileNet, which is due mainly to the substantially larger model size of MobileNet-L.

\subsection{Hyperparameter Sensitivity}

There are two main hyperparameters, namely $T$ and $\lambda$, that influence the performance of HDUS. In this section, we study the effect of different hyperparameter values to our proposed framework in the heterogeneous setting.

$T$ denotes the temperature in knowledge distillation, which can soften the logits produced by the softmax function in local main models (teacher models). 
$T=1$ means the normal softmax function is applied. $\lambda \in [0, 1)$ is a trade-off coefficient that balances the contribution of local model and seed models in each client. Higher $\lambda$ means the ensemble relies more on the knowledge from neighbors. 
In general, the results in Fig.\ref{fig:Hyper} illustrate that the magnifying effect from $\lambda$ is more obvious, and MobileNet generally performs better than ResNet in FMNIST and Cifar10 under a wide range of parameter settings. 

\section{Conclusion}

One main obstacle in designing a decentralized unlearning framework is that the unlearning requests send after the knowledge is shared. In this case, some clients may fail to erase their information from all involved clients since the medium clients who built the connections may have already left the network and hence suspend the unlearning requests. On the other hand, clients in conventional decentralized frameworks blend with peer knowledge via training. This means restoring the model to the state right before collaboration is not significantly faster than retraining the model from the initial state.  

To this end, we present HDUS, a machine unlearning framework for heterogeneous decentralized collaboration. In our methods, we assume the central server in conventional unlearning frameworks is absent and allows each client (client) to possess seed models from its neighbors. In this way, each client operates as an ensemble that combines the knowledge from the local dataset and neighbors' datasets. Since the seed models are trained via an independent reference dataset, each client cannot deduce the information of their neighbors' local data. Additionally, since each client only sends the seed model of its local main model to its neighbors, it avoids knowledge fusion among the network. Otherwise, if one client proposes an unlearn request, all other clients have to modify their models or ensembles, which is impractical in large-scale networks with frequent unlearn requests.

To the best of our knowledge, this is the first work that solves the unlearning problem in decentralized collaboration scenarios. The proposed easy-to-erase ensemble architecture not only speeds up the unlearning process and raises the unlearning request tolerance of the whole decentralized network but also enhances its compatibility with different task settings (e.g., dynamic linkages). Nevertheless, HDUS has the following limitations. By design, the knowledge of one client cannot be spread to the whole network, which suppresses the advantage of inter-device communications. Besides, the weights on different sub-models are set as equal and the connections between devices are static. We plan to introduce a semi-decentralized framework in the future to coordinate the clients with a dynamic neighbor allocation, which may further improve the network performance.

\section*{Acknowledgments}
This work is supported by the Australian Research Council under the streams of Future Fellowship (No. FT210100624), Discovery Project (No. DP190101985), and Discovery Early Career Researcher Award (No. DE230101033).

\section*{Conflict of interest}
The authors declare no potential conflict of interests.

\section*{Data Availability Statement}
This paper is evaluated with publicly available benchmarks, which can be accessed via their references.

\bibliography{references}

\end{document}